\documentclass[runningheads]{llncs}
\usepackage{graphicx,subfig,float}
\usepackage[table,xcdraw]{xcolor}

\usepackage{blindtext}

\usepackage[utf8]{inputenc} 
\usepackage[T1]{fontenc}    
\usepackage{hyperref}       
\usepackage{url}            
\usepackage{booktabs}       
\usepackage{amsfonts}       
\usepackage{nicefrac}       
\usepackage{microtype}      
\usepackage{amsfonts}       
\usepackage{nicefrac}       
\usepackage{microtype}      
\usepackage{amsmath}
\usepackage{color,mathtools}
\usepackage{enumitem}
\usepackage{multirow}
\usepackage{graphicx}
\usepackage{bm,bbm}
\usepackage{rotating}
\usepackage{pgffor}
\usepackage{array}
\usepackage{tabularx}

\newcommand{\punt}[1]{}

\newcommand{\vx}{\mathbf{x}}

\newcommand{\vz}{\mathbf{z}}

\newcommand{\bq}{\begin{equation}}
\newcommand{\eq}{\end{equation}}
\newcommand{\ba}{\begin{eqnarray}}
\newcommand{\ea}{\end{eqnarray}}

\newcommand{\remove}[1]{}

\begin{document}
%
%
%
\title{Calibrating Healthcare AI: Towards Reliable and Interpretable Deep Predictive Models}
\titlerunning{Towards Reliable and Interpretable Deep Predictive Models}
\author{Jayaraman J. Thiagarajan$^+$\thanks{This work was performed under the auspices of the U.S. Department of Energy by Lawrence Livermore National Laboratory under Contract DE-AC52-07NA27344.}, Prasanna Sattigeri$^{\dagger}$, \\ Deepta Rajan$^{\dagger}$, Bindya Venkatesh$^{\ddagger}$}
\institute{$^{+}$Lawrence Livermore National Labs, $^{\dagger}$IBM Research AI, $^{\ddagger}$Arizona State University}
\maketitle              
\begin{abstract}
The wide-spread adoption of representation learning technologies in clinical decision making strongly emphasizes the need for characterizing model reliability and enabling rigorous introspection of model behavior. While the former need is often addressed by incorporating uncertainty quantification strategies, the latter challenge is addressed using a broad class of interpretability techniques. In this paper, we argue that these two objectives are not necessarily disparate and propose to utilize prediction calibration to meet both objectives. More specifically, our approach is comprised of a calibration-driven learning method, which is also used to design an interpretability technique based on counterfactual reasoning. Furthermore, we introduce \textit{reliability plots}, a holistic evaluation mechanism for model reliability. Using a lesion classification problem with dermoscopy images, we demonstrate the effectiveness of our approach and infer interesting insights about the model behavior. 

\end{abstract}

\section{Motivation}
Artificial intelligence (AI) techniques such as deep learning have achieved unprecedented success with critical decision-making, from diagnosing diseases to prescribing treatments, in healthcare~\cite{faust2018deep,kononenko2001machine,miotto2018deep}. However, to prioritize patient safety, one must ensure such methods are accurate and reliable~\cite{ching2018opportunities}. For example, a neural network model can produce highly concentrated softmax probabilities – suggesting a reliable class assignment – even for out-of-distribution test samples, which indicates that the confidences are not well-calibrated. This strongly emphasizes the need to both reliably assess model's confidences, and enable rigorous introspection of model behavior~\cite{cabitza2019wants,ching2018opportunities,tonekaboni2019clinicians}. While the former objective can be handled by incorporating a variety of prediction calibration strategies~\cite{guo2017calibration,leibig2017leveraging}, a large class of interpretability tools are used to support model introspection~\cite{ahmad2018interpretable,montavon2018methods,arya2019one,guidotti2018survey}. 

Broadly, prediction calibration is the process of adjusting predictions to improve the error distribution of a predictive model -- example range from data augmentation techniques~\cite{thulasidasan2019mixup} and regularization strategies~\cite{seo2019learning} to more sophisticated methods that quantify the \textit{epistemic} (or model) uncertainties and \textit{aleatoric} (or data) uncertainties for calibrating model confidences~\cite{gal2016uncertainty,ghahramani2015probabilistic,kuleshov2015calibrated,tagasovska2018frequentist}. On the other hand, widely adopted interpretability tools have focused on accumulating evidences (e.g. saliency maps or feature importances) to explain local (single sample) \cite{lime,unifiedPI} or global (groups of samples or the entire dataset) behavior of a trained model \cite{nguyen2016multifaceted,deepling}. In this paper, we hypothesize that the two objectives of improving model reliability and enabling rigorous introspection are not necessarily disparate and that prediction calibration can be used to achieve both.

\noindent \textbf{Contributions.} First, we propose a novel calibration-driven learning approach, which produces prediction intervals for each image instead of point estimates, and utilizes an interval calibration objective to learn the model parameters. Second, we introduce \textit{reliability plots}, which quantify the trade-off between model autonomy and improved generalization by including experts in the loop during inference, as a holistic evaluation mechanism of model reliability. Third and more importantly, we develop a novel interpretability technique that enables us to rigorously explore model behavior (local) via counterfactual evidences generated in a disentangled latent space through prediction calibration.

\noindent \textbf{Findings.} We use a lesion classification problem with dermoscopy images to evaluate the proposed methods. Using both conventional metrics, as well as our \textit{reliability plots}, we find that our approach produces superior models when compared to commonly adopted solutions, including deep networks and ensembling methods. Finally, using the proposed interpretability technique, we make a number of key findings about the model behavior, that would not be apparent otherwise - our findings include spurious correlations, intricate relationships between different classes, regimes of uncertainty and a comprehensive understanding of model strengths and weaknesses. Together, the proposed methods provide a completely new way to build and analyze models in healthcare applications.

\section{Dataset and Problem Description}
In this paper, we use the ISIC 2018 lesion diagnosis challenge dataset~\cite{codella2019skin,tschandl2018ham10000}, which contains a total of $10,015$ dermoscopic lesion images (corresponding to the labeled training set) from the HAM10000 database~\cite{tschandl2018ham10000}. The images were acquired with a variety of dermatoscope types from a historical sample of patients presented for skin cancer screening from multiple institutions. Each image is associated with one out of $7$ disease states: Melanoma, Melanocytic nevus, Basal cell carcinoma, Actinic keratosis, Benign keratosis, Dermatofibroma and Vascular lesion. The goal is to build a classifier to predict the disease type from the image, while satisfying the key design objectives of improved model reliability and being interpretable. Dermatologists use rules of thumb when initially investigating a skin lesion, for example the widely adopted ABCD signatures: asymmetry, border, color, and diameter. This naturally motivates the use of representation learning approaches that can automatically infer latent concepts to effectively describe the distribution of images in different classes.


%

\section{Calibration-Driven Predictive Modeling}

\subsection{Disentangled Latent Representations}
Supervised models built upon representations that align well with true generative factors of data have been found to be robust and interpretable. Most real-world problems involve raw observations without any supervision about the generative factors. Consequently, the use of latent generative models with \textit{disentanglement} has become popular, wherein the goal is to recover a latent space with statistically independent dimensions. A small change in one of the dimensions of such representations often produces interpretable change in the generated data sample.

\begin{figure}[t]
    \centering
    \includegraphics[width=0.99\linewidth]{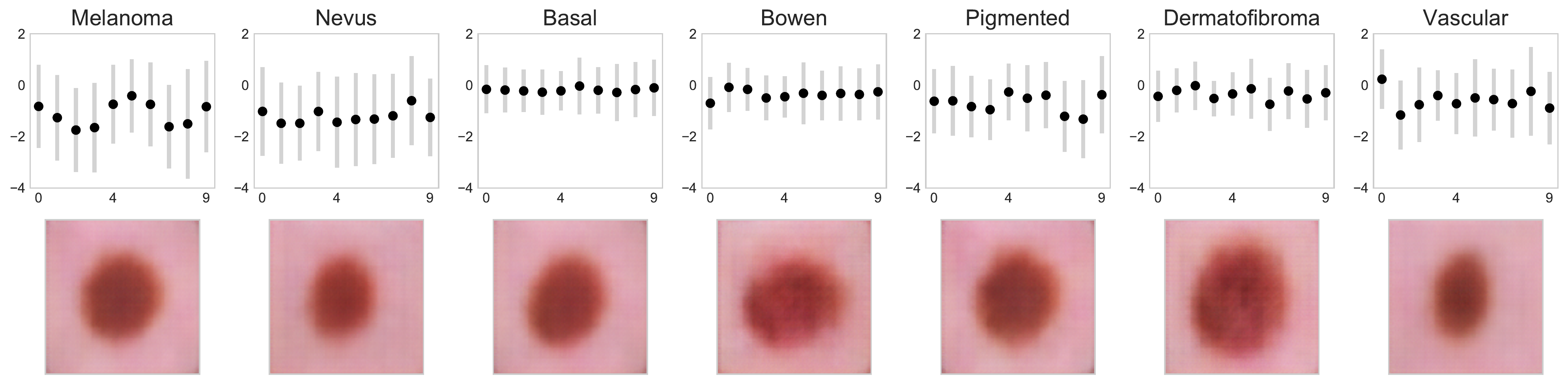}
    \caption{Distribution of latent features in each class obtained using DIP-VAE~\cite{kumar2017variational}.}
    \label{fig:latent}
\end{figure}

In our approach, we use DIP-VAE~\cite{kumar2017variational}, a variant of Variational Autoencoders (VAE), which has been shown to be effective on standard disentanglement benchmarks. The conventional VAE works with a relatively simple and disentangled prior $p(\mathbf{z})$ with no explicit interaction among the latent dimensions (e.g., the standard normal $\mathcal{N}(0,I)$). The complexity of the observed data $\mathbf{x}$, modeled by the decoder, is absorbed in the conditional distribution $p(\mathbf{x}|\mathbf{z})$ which infers the interactions among latent dimensions. Even though the prior is disentangled, it is possible that the variational distribution $q(\vz) = \int q(\vz|\vx) p(\vx) d\vx$ (\textit{aggregated-posterior}), induced over the latent space, modeled by the encoder, is not disentangled.  DIP-VAE encourages a disentangled aggregated-posterior by matching the covariance of the two distributions $q(\vz)$ and $p(\vz)$. This amounts to decorrelating the dimensions of the inferred latent space. Figure~\ref{fig:latent} shows the distribution of latent features obtained using DIP-VAE (10 latent dimensions) for each of the $7$ classes. We also show the decoder reconstruction for the average latent representation in each class.



\subsection{Learning Deep Models via Interval Calibration}
Our approach utilizes the notion of interval calibration~\cite{thiagarajan2019learn} to design predictive models. First, we begin by assuming that our model produces prediction intervals instead of simple point estimates, i.e., $[\hat{\mathbf{y}} - \boldsymbol{\delta}, \hat{\mathbf{y}} + \boldsymbol{\delta}]$, for each image $\mathbf{x}$ from its latent representation $\mathbf{z}$. Our model is comprised of two modules $f$ and $g$, implemented as neural networks, to produce estimates $\hat{\mathbf{y}} = f(\mathbf{z})$ and $\boldsymbol{\delta} = g(\mathbf{z})$ respectively. Here, $\hat{\mathbf{y}} \in \mathbb{R}^K$ is a vector of predicted logits for each of the $K$ classes. Since we operate on the logits directly, we also transform the ground truth labels into logits. In practice, we found that smoothing the labels before converting them into logits led to improved convergence. For example, a sample belonging to class $1$ was assigned logits $[+1, -2, -2, -2, -2, -2, -2]$, which allows a small non-zero probability ($\approx 0.04$) to the negative classes. Suppose that the likelihood for the true $\mathbf{y}$ to be contained in the interval is $p(\hat{\mathbf{y}} - \boldsymbol{\delta} \leq \mathbf{y}  \leq \hat{\mathbf{y}} + \boldsymbol{\delta})$, the intervals are considered to be well calibrated if the likelihood matches the expected confidence level. For a confidence level $\alpha$, we expect the interval to contain the true target for $100\times \alpha\%$ of the samples from the unknown distribution $p(\mathbf{x})$. 

We design an alternating optimization strategy to infer $\mathrm{\theta}$ and $\mathrm{\phi}$, parameters of models $f$ and $g$ respectively, using labeled data $\{(\mathbf{x}_i,\mathbf{y}_i)\}_{i=1}^N$. In order to update the parameters of $g$, we use an empirical interval calibration error, similar to~\cite{thiagarajan2019building}, evaluated using mini-batches:
\begin{align}
    \mathrm{\phi}^* = \arg \min_{\mathrm{\phi}} &\sum_{k=1}^K \left|\alpha - 
    \frac{1}{N} \sum_{i=1}^N \mathbbm{1}\bigg[(\hat{\mathbf{y}}_i[k] - \boldsymbol{\delta}_i[k])  \leq \mathbf{y}_i[k]  \leq (\hat{\mathbf{y}}_i[k] + \boldsymbol{\delta}_i[k])\bigg] \right|, 
    \label{eqn:calib}
\end{align}where $\boldsymbol{\delta}_i = g(\mathbf{z}_i;\mathrm{\phi})$, $\mathbf{y}_i[k]$ is the $k^{\text{th}}$ element of the vector $\mathbf{y}_i$ and the desired confidence level $\alpha$ is an input to the algorithm. When updating the parameters $\mathrm{\phi}$, we assume that the estimator $f(.;\mathrm{\theta})$ is known and fixed. Now, given the updated $\mathrm{\phi}$, we learn the parameters $\mathrm{\theta}$ using the following hinge-loss objective:
\begin{align}
    \nonumber \mathrm{\theta}^* = \arg \min_{\mathrm{\theta}}  \sum_{k=1}^K \frac{1}{N} \sum_{i = 1}^N \bigg[&\max\bigg(0, (\hat{\mathbf{y}}_i[k] - \boldsymbol{\delta}_i[k]) - \mathbf{y}_i[k] +\tau\bigg) + \\
    &\max\bigg(0, \mathbf{y}_i[k] - (\hat{\mathbf{y}}_i[k] + \boldsymbol{\delta}_i[k]) +\tau\bigg) \bigg],
    \label{eqn:hinge}
\end{align}where $\hat{\mathbf{y}}_i = f(\mathbf{z}; \mathrm{\theta})$ and $\tau$ is a threshold set to $0.05$ in our experiments.  Intuitively, for a fixed $\mathrm{\phi}$, obtaining improved estimates for $\hat{\mathbf{y}}$ can increase the empirical calibration error in \eqref{eqn:calib} by achieving higher likelihoods even for lower confidence levels.  However, in the subsequent step of updating $\mathrm{\phi}$, we expect $\boldsymbol{\delta}^{'}$s to become sharper in order to reduce the calibration error. This collaborative optimization process thus leads to superior quality point estimates and highly calibrated intervals. We repeat the two steps (eqns. \eqref{eqn:calib} and \eqref{eqn:hinge}) until convergence. In our experiments, we set the desired confidence level $\alpha = 0.7$. Further, both $f$ and $g$ were designed as $5-$layer fully connected networks with hidden sizes $[64,128,256,64,7]$ and \textit{ReLU} activations. We use the Adam optimizier with learning rates $3e-4$ and $1e-4$ for the two models.

\begin{figure}[t]
    \centering
    \includegraphics[width=0.99\linewidth,keepaspectratio]{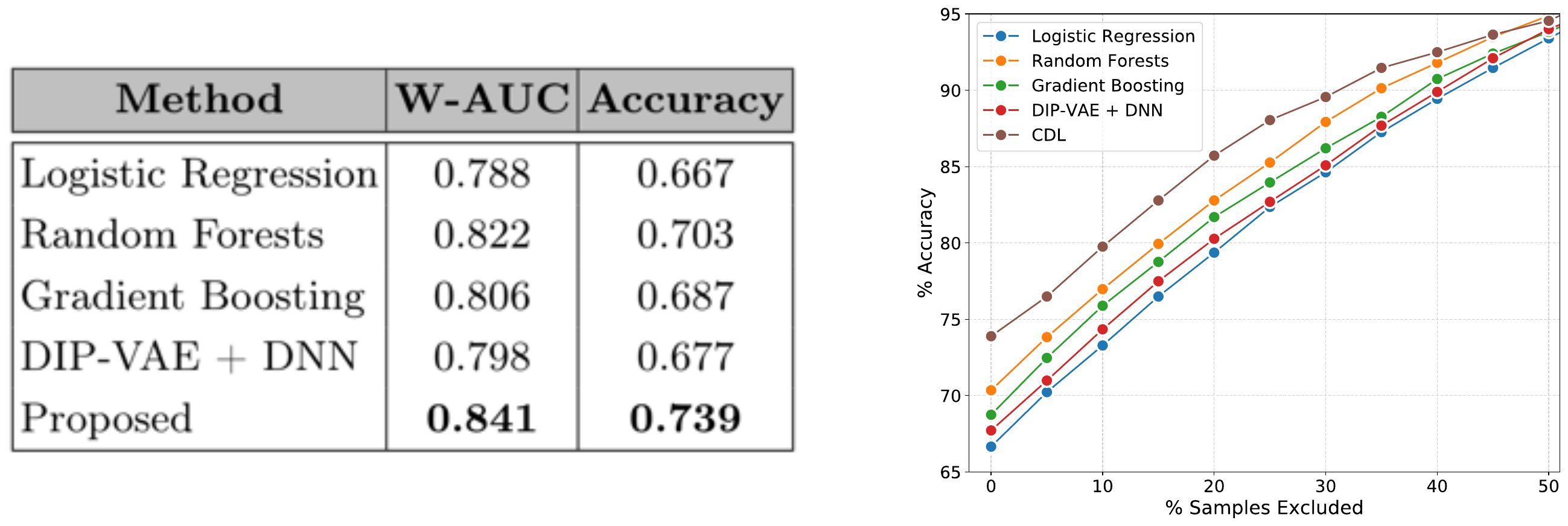} 
    \caption{Performance Evaluation - Comparing prediction performance averaged from 3-fold cross validation (left); Reliability plots for different approaches (right).}
    \label{fig:perf-cdl}
\end{figure}

\subsection{Evaluating Model Reliability}
While metrics such as accuracy and area under ROC have been widely adopted for evaluating model performance, we argue that it is critical to understand how calibrated the confidences of a model are, in order to quantify its reliability. In particular, we study the trade-off between model autonomy and expected test-time performance by including experts in the loop during inference. We use the held-out validation set to construct a \textit{reliability plot} as follows: We first measure the model's confidence on a prediction for each sample using the entropy of the softmax probabilities, $\mathcal{H}(\boldsymbol{\rho}) = \sum_{k=1}^K -\boldsymbol{\rho}[k] \log \boldsymbol{\rho}[k].$ where $\boldsymbol{\rho} = \texttt{Softmax}(\hat{\mathbf{y}})$. Subsequently, we rank the samples based on their confidences, and hypothesize that one can use the model's predictions for the most confident cases and engage the expert to label less confident samples (i.e. use the true labels from the validation set). The overall performance is obtained by combining the predictions from both the model and the expert. In an ideal scenario, one would expect high validation accuracies for the model, while requiring minimal expert involvement. A \textit{reliability plot} summarizes this trade-off by varying the $\%$ Samples deferred by the model to an expert and measuring the validation accuracy in each case.


\begin{figure}[t]
    \centering
    \includegraphics[width=0.99\linewidth]{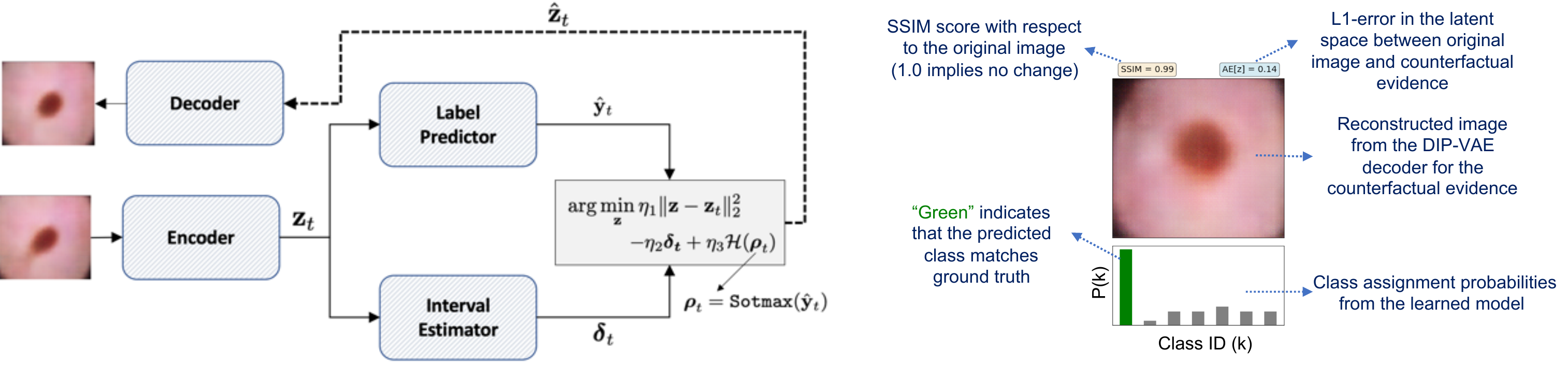}
    \caption{Proposed approach for model introspection - Illustration of the optimization strategy for generating counterfactual evidences (left); Description of components in the visual layout used for showing our results.}
    \label{fig:layout}
\end{figure}

\subsection{Empirical Results}
In Figure \ref{fig:perf-cdl}, we report the average performance from $3-$fold cross validation of the proposed approach, in comparison to different popular baselines including logistic regression, random forests, gradient boosting and a deep network with the cross-entropy loss. Note, we used only the labeled training set from the ISIC 2018 dataset for our evaluation and all baselines were trained with the disentangled latent representations from DIP-VAE. The results clearly show that the proposed approach consistently outperforms the baselines, in terms of conventional evaluation metrics such as weighted-AUC and macro accuracy, as well as the proposed reliability plots. More specifically, when compared to standard cross-entropy based training, calibration-driven learning produces more reliable models. For example, our approach achieves $80\%$ accuracy on this challenging benchmark with only $10\%$ samples being deferred to the expert, in contrast to the $74\%$ accuracy of the standard neural network.

\section{Model Introspection via Counterfactual Reasoning }
An important hypothesis of this paper is that prediction calibration can elucidate the behavior of a trained model. In addition to enabling practitioners build trust on AI systems, our introspection approach can shed light on strengths and weaknesses of the model. While there has been considerable effort in quantifying uncertainties in machine learning models and presenting users with expected variability in the predictions~\cite{gal2016uncertainty,ghahramani2015probabilistic}, they are not human interpretable unless they can be mapped to patterns in the input data. Hence, we propose to generate counterfactual evidences for a given sample, through exploration in the disentangled latent space, which enable users to quickly grasp the regimes of confidence and uncertainty.

Figure~\ref{fig:layout}(left) illustrates the proposed approach. Given a test image $\mathbf{x}_t$, we first compute its latent representation $\mathbf{z}_t$ using the DIP-VAE encoder. We then use the pre-trained models $f$ (label predictor) and $g$ (interval estimator) from the proposed calibration-driven learning approach to invoke our exploratory analysis. In general, counterfactual reasoning refers to the process of identifying alternative possibilities of events that could lead to completely different outcomes~\cite{schulam2017reliable}. For example, adversarial attacks are routinely designed in machine learning by identifying imperceptible image perturbations that can fool a pre-trained classifier. For the first time, we show that counterfactual reasoning can be effectively utilized to introspect models by enabling predictions with varying levels of confidence, and performing this optimization in the disentangled latent space will ensure that the counterfactual examples are indeed physically plausible. We propose the following inference-time optimization to generate counterfactual evidences:
\begin{equation}
    \hat{\mathbf{z}}_t = \arg \min_{\mathbf{z}} \eta_1 \|\mathbf{z}_t - \mathbf{z}\|_2^2 - \eta_2 g(\mathbf{z}; \mathrm{\phi}) + \eta_3 \mathcal{H}(\boldsymbol{\rho}), \text{where } \boldsymbol{\rho} = \texttt{Softmax}(f(\mathbf{z};\mathrm{\theta})).
    \label{eqn:int}
\end{equation}Here $\eta_1, \eta_2, \eta_3$ are user-defined hyper-parameters. The first term ensures that the generated evidence is not semantically different from the given image. The second term attempts to increase the interval width to improve the likelihood of the true prediction to be contained in the interval. Finally, the third term directly controls the confidence of the prediction (in terms of entropy). In essence, this optimization searches for an evidence in the latent space that is semantically similar to a given image, likely to be well calibrated and can produce prediction probabilities with low entropy (more confident). Optionally, one can change the sign of the third term and search for evidences with high entropy (less confident). In our analysis, we fixed $\eta_2 = 0.5$ and $\eta_3 = 0.2$, and varied $\eta_1$ to generate evidences with increasing amounts of disparity with respect to the given image.

\begin{figure}[t]
    \centering
    \includegraphics[width=0.95\linewidth]{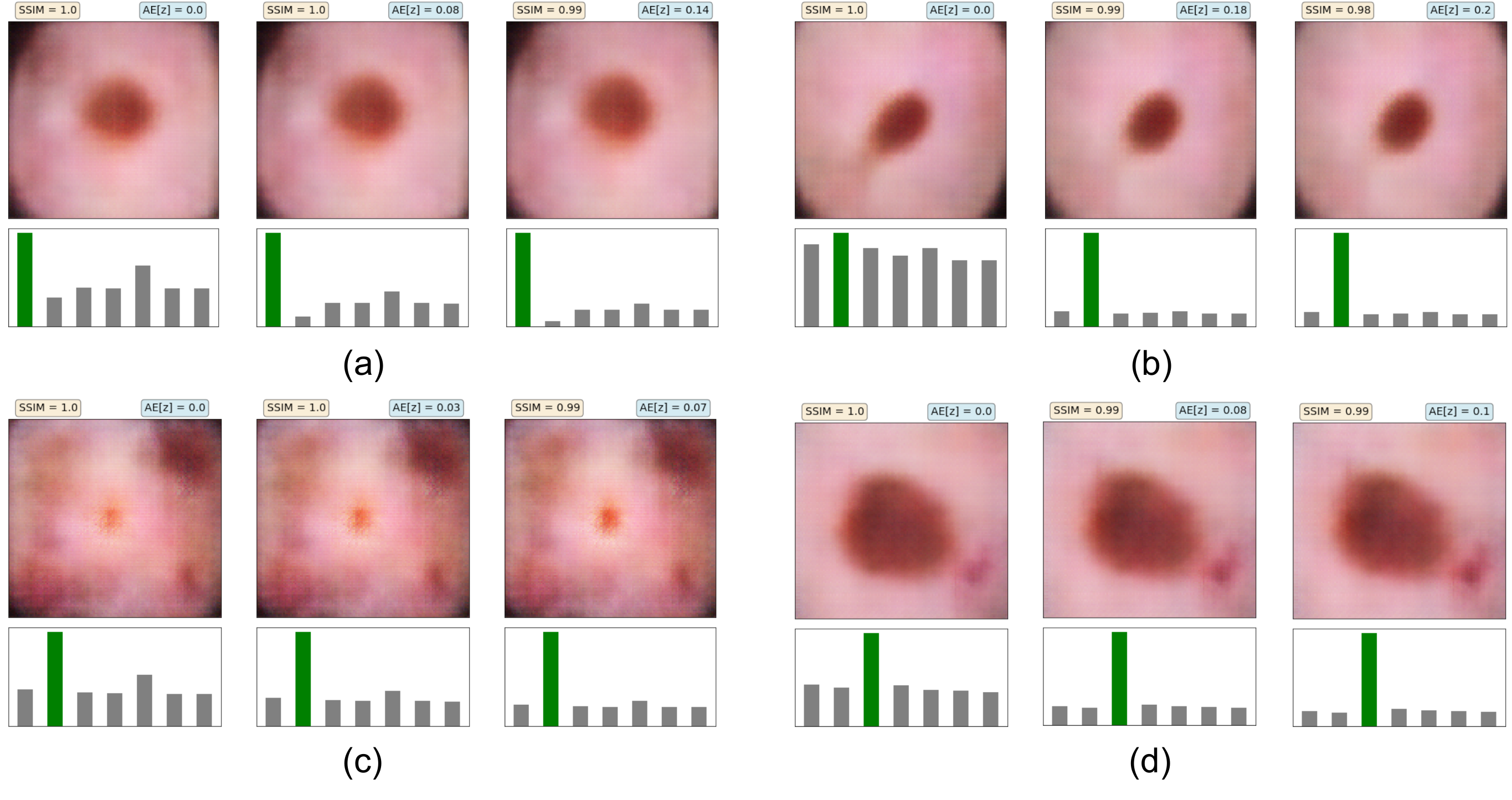}
    \caption{Counterfactual evidences with low semantic discrepancy (as shown by the high SSIM scores) but significantly higher confidence predictions.}
    \label{fig:low_ent}
\end{figure}

\begin{figure}[t]
    \centering
    \includegraphics[width=0.95\linewidth]{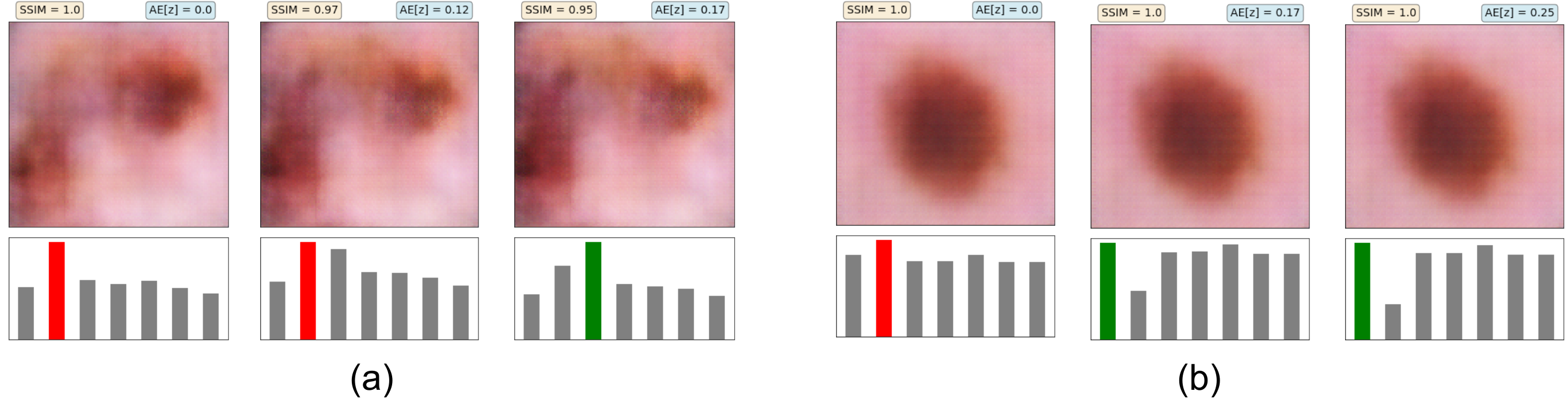}
    \caption{Using the proposed approach, we can identify how subtle differences in image statistics can produce very different disease outcomes.}
    \label{fig:class_conf}
\end{figure}

Figure~\ref{fig:layout}(right) describes the components of the visual layout that we use to show results from our analysis. While the top row shows the generated evidence (using the DIP-VAE decoder on $\hat{\mathbf{z}}$), the bottom row shows its softmax probabilities ($7$ classes) from the label predictor. Since we have access to the true labels for the held-out validation set, we indicate the predicted class in \textit{green} when it matches the ground truth and in \textit{red} otherwise. Furthermore, we show the discrepancy between the given image and the evidence in the latent space, via the average $\ell_1$ error AE($\mathbf{z}$), and in the image space, via the structural similarity metric (SSIM)~\cite{wang2004image}.  

\begin{figure}[t]
    \centering
    \includegraphics[width=0.95\linewidth]{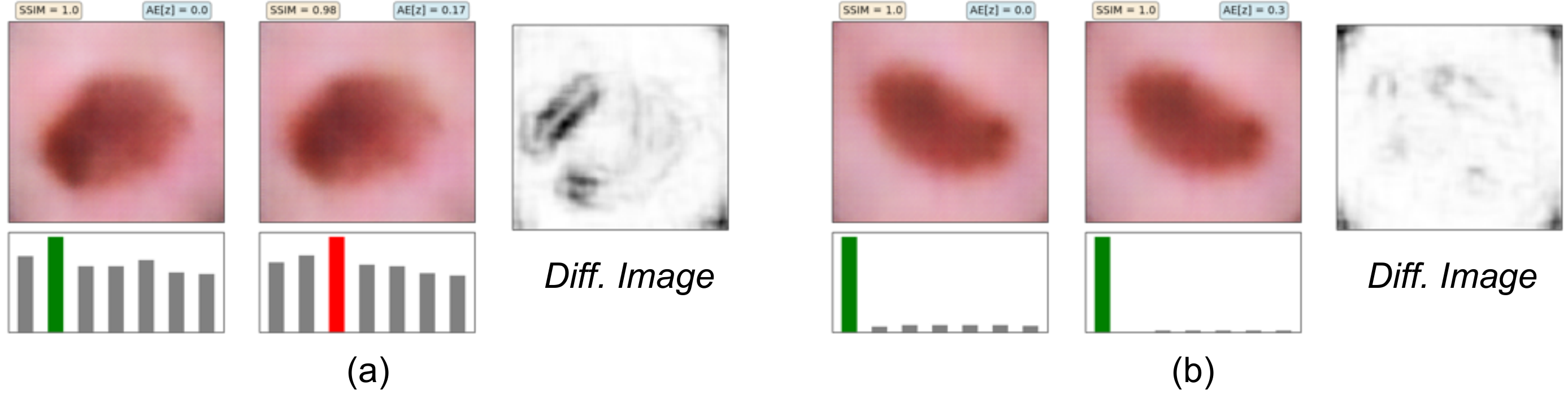}
    \caption{We find that even a well-performing classifier picks up spurious correlations by responding to the irrelevant corner regions along with the actual lesions.}
    \label{fig:spurious}
\end{figure}

\begin{figure}[t]
    \centering
    \includegraphics[width=0.95\linewidth]{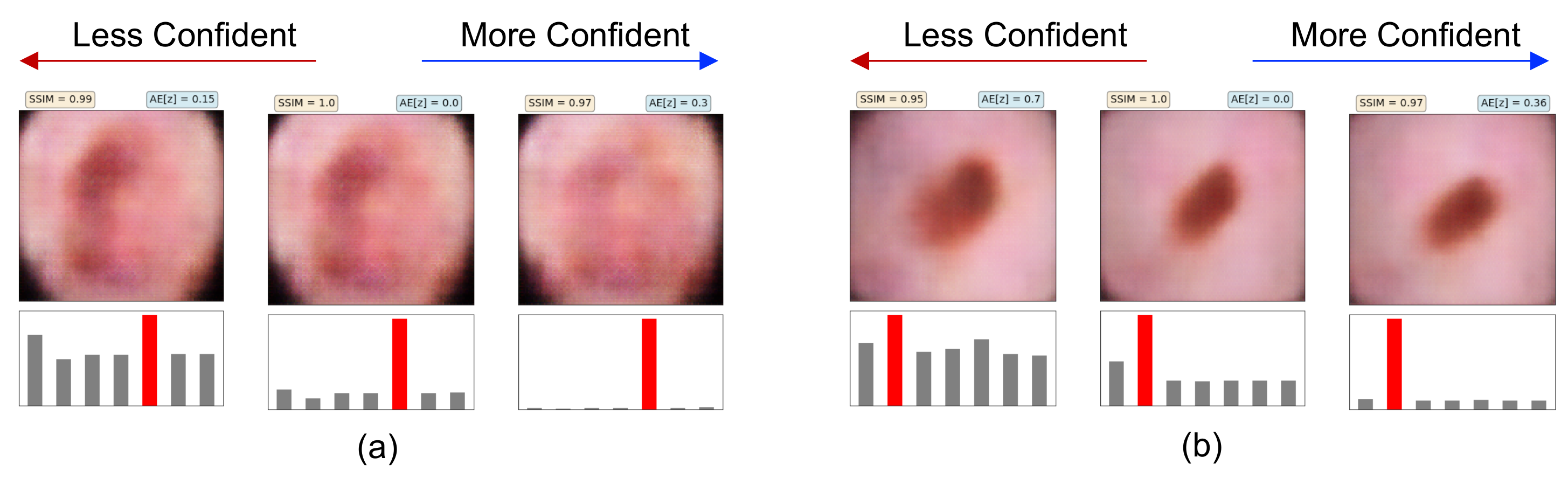}
    \caption{By simultaneously viewing the evidences in different confidence regimes, one can obtain a holistic understanding of the model.}
    \label{fig:regime}
\end{figure}

\noindent \textbf{Key Findings.} Using the proposed introspection approach, we made a number of interesting observations about the lesion classification model: (i) As showed in Figure \ref{fig:low_ent}, we are able to generate \textbf{counterfactual evidences with low semantic discrepancy but significantly lower entropy}. This indicates that the model has inferred statistical patterns specific to each of the classes, which can be emphasized to produce highly confident predictions. This includes adding patterns that are correlated with the disease condition (see Figure~\ref{fig:low_ent}(c) where the lesion region intensities are emphasized) or removing patterns that are uncorrelated to the suspected condition (see Figure~\ref{fig:low_ent}(b) where the apparent tail pattern is leading to highly uncertain prediction); (ii) We are able to \textbf{identify intricate relationships between classes through evidences} (see Figure \ref{fig:class_conf}). Through the inference-time calibration, we are able to generate evidences with very different disease outcomes for subtle differences in the image statistics. For example in Figure \ref{fig:class_conf}(b), we are able to completely eliminate the possibility of \textit{Melanocytic nevus} through minimal movement in the disentangled latent space; (iii) Our analysis shows that \textbf{even a well-performing classifier can still pick up spurious correlations} and identifying them is essential for enabling model trust. As showed in Figure~\ref{fig:spurious}, the model can produce different outcomes (Figure~\ref{fig:spurious}(a)) or more confident predictions (Figure~\ref{fig:spurious}(b)) by relying on irrelevant corner regions, in addition to the actual lesion pixels; (iv) Finally, by generating evidences for different confidence regimes (by changing the sign of the entropy term in eqn. \eqref{eqn:int}), one can \textbf{understand how different image patterns contribute to increasing or decreasing the prediction confidences}. As showed in Figure~\ref{fig:regime}, easily interpretable factors such as asymmetry, border, color and diameter can be used to analyze the characteristics of evidences in different confidence regimes and obtain a holistic understanding of the model's behavior. 

In summary, prediction calibration is an effective principle for designing reliable models as well as building tools for rigorous model introspection. Our analysis with the lesion classification dataset clearly demonstrates the different kinds of insights one can infer by performing counterfactual reasoning via prediction calibration with disentangled latent spaces.

\bibliographystyle{splncs04}
\bibliography{refs}

\begin{thebibliography}{10}
\providecommand{\url}[1]{\texttt{#1}}
\providecommand{\urlprefix}{URL }
\providecommand{\doi}[1]{https://doi.org/#1}

\bibitem{ahmad2018interpretable}
Ahmad, M.A., Eckert, C., Teredesai, A.: Interpretable machine learning in
  healthcare. In: Proceedings of the 2018 ACM International Conference on
  Bioinformatics, Computational Biology, and Health Informatics. pp. 559--560
  (2018)

\bibitem{arya2019one}
Arya, V., Bellamy, R.K., Chen, P.Y., Dhurandhar, A., Hind, M., Hoffman, S.C.,
  Houde, S., Liao, Q.V., Luss, R., Mojsilovi{\'c}, A., et~al.: One explanation
  does not fit all: A toolkit and taxonomy of ai explainability techniques.
  arXiv preprint arXiv:1909.03012  (2019)

\bibitem{cabitza2019wants}
Cabitza, F., Campagner, A.: Who wants accurate models? arguing for a different
  metrics to take classification models seriously. arXiv preprint
  arXiv:1910.09246  (2019)

\bibitem{ching2018opportunities}
Ching, T., Himmelstein, D.S., Beaulieu-Jones, B.K., Kalinin, A.A., Do, B.T.,
  Way, G.P., Ferrero, E., Agapow, P.M., Zietz, M., Hoffman, M.M., et~al.:
  Opportunities and obstacles for deep learning in biology and medicine.
  Journal of The Royal Society Interface  \textbf{15}(141),  20170387 (2018)

\bibitem{codella2019skin}
Codella, N., Rotemberg, V., Tschandl, P., Celebi, M.E., Dusza, S., Gutman, D.,
  Helba, B., Kalloo, A., Liopyris, K., Marchetti, M., et~al.: Skin lesion
  analysis toward melanoma detection 2018: A challenge hosted by the
  international skin imaging collaboration (isic). arXiv preprint
  arXiv:1902.03368  (2019)

\bibitem{faust2018deep}
Faust, O., Hagiwara, Y., Hong, T.J., Lih, O.S., Acharya, U.R.: Deep learning
  for healthcare applications based on physiological signals: A review.
  Computer methods and programs in biomedicine  \textbf{161},  1--13 (2018)

\bibitem{gal2016uncertainty}
Gal, Y.: Uncertainty in deep learning. University of Cambridge  \textbf{1}, ~3
  (2016)

\bibitem{ghahramani2015probabilistic}
Ghahramani, Z.: Probabilistic machine learning and artificial intelligence.
  Nature  \textbf{521}(7553),  452--459 (2015)

\bibitem{guidotti2018survey}
Guidotti, R., Monreale, A., Ruggieri, S., Turini, F., Giannotti, F., Pedreschi,
  D.: A survey of methods for explaining black box models. ACM computing
  surveys (CSUR)  \textbf{51}(5),  1--42 (2018)

\bibitem{guo2017calibration}
Guo, C., Pleiss, G., Sun, Y., Weinberger, K.Q.: On calibration of modern neural
  networks. In: Proceedings of the 34th International Conference on Machine
  Learning-Volume 70. pp. 1321--1330. JMLR. org (2017)

\bibitem{kononenko2001machine}
Kononenko, I.: Machine learning for medical diagnosis: history, state of the
  art and perspective. Artificial Intelligence in medicine  \textbf{23}(1),
  89--109 (2001)

\bibitem{kuleshov2015calibrated}
Kuleshov, V., Liang, P.S.: Calibrated structured prediction. In: Advances in
  Neural Information Processing Systems. pp. 3474--3482 (2015)

\bibitem{kumar2017variational}
Kumar, A., Sattigeri, P., Balakrishnan, A.: Variational inference of
  disentangled latent concepts from unlabeled observations. arXiv preprint
  arXiv:1711.00848  (2017)

\bibitem{leibig2017leveraging}
Leibig, C., Allken, V., Ayhan, M.S., Berens, P., Wahl, S.: Leveraging
  uncertainty information from deep neural networks for disease detection.
  Scientific reports  \textbf{7}(1),  1--14 (2017)

\bibitem{unifiedPI}
Lundberg, S., Lee, S.I.: Unified framework for interpretable methods. In:
  Advances of Neural Inf. Proc. Systems (2017)

\bibitem{miotto2018deep}
Miotto, R., Wang, F., Wang, S., Jiang, X., Dudley, J.T.: Deep learning for
  healthcare: review, opportunities and challenges. Briefings in bioinformatics
   \textbf{19}(6),  1236--1246 (2018)

\bibitem{montavon2018methods}
Montavon, G., Samek, W., M{\"u}ller, K.R.: Methods for interpreting and
  understanding deep neural networks. Digital Signal Processing  \textbf{73},
  1--15 (2018)

\bibitem{nguyen2016multifaceted}
Nguyen, A., Yosinski, J., Clune, J.: Multifaceted feature visualization:
  Uncovering the different types of features learned by each neuron in deep
  neural networks. arXiv preprint arXiv:1602.03616  (2016)

\bibitem{lime}
Ribeiro, M., Singh, S., Guestrin, C.: ``{Why} should {I} trust you?''
  {Explaining} the predictions of any classifier. In: ACM SIGKDD Intl.
  Conference on Knowledge Discovery and Data Mining (2016)

\bibitem{schulam2017reliable}
Schulam, P., Saria, S.: Reliable decision support using counterfactual models.
  In: Advances in Neural Information Processing Systems. pp. 1697--1708 (2017)

\bibitem{seo2019learning}
Seo, S., Seo, P.H., Han, B.: Learning for single-shot confidence calibration in
  deep neural networks through stochastic inferences. In: Proceedings of the
  IEEE Conference on Computer Vision and Pattern Recognition. pp. 9030--9038
  (2019)

\bibitem{tagasovska2018frequentist}
Tagasovska, N., Lopez-Paz, D.: Frequentist uncertainty estimates for deep
  learning. arXiv preprint arXiv:1811.00908  (2018)

\bibitem{thiagarajan2019learn}
Thiagarajan, J.J., Venkatesh, B., Rajan, D.: Learn-by-calibrating: Using
  calibration as a training objective. arXiv preprint arXiv:1910.14175  (2019)

\bibitem{thiagarajan2019building}
Thiagarajan, J.J., Venkatesh, B., Sattigeri, P., Bremer, P.T.: Building
  calibrated deep models via uncertainty matching with auxiliary interval
  predictors. AAAI Conference on Artificial Intelligence  (2019)

\bibitem{thulasidasan2019mixup}
Thulasidasan, S., Chennupati, G., Bilmes, J.A., Bhattacharya, T., Michalak, S.:
  On mixup training: Improved calibration and predictive uncertainty for deep
  neural networks. In: Advances in Neural Information Processing Systems. pp.
  13888--13899 (2019)

\bibitem{tonekaboni2019clinicians}
Tonekaboni, S., Joshi, S., McCradden, M.D., Goldenberg, A.: What clinicians
  want: contextualizing explainable machine learning for clinical end use.
  arXiv preprint arXiv:1905.05134  (2019)

\bibitem{tschandl2018ham10000}
Tschandl, P., Rosendahl, C., Kittler, H.: The ham10000 dataset, a large
  collection of multi-source dermatoscopic images of common pigmented skin
  lesions. Scientific data  \textbf{5},  180161 (2018)

\bibitem{wang2004image}
Wang, Z., Bovik, A.C., Sheikh, H.R., Simoncelli, E.P.: Image quality
  assessment: from error visibility to structural similarity. IEEE transactions
  on image processing  \textbf{13}(4),  600--612 (2004)

\bibitem{deepling}
Weidele, D., Strobelt, H., Martino, M.: Deepling: A visual interpretability
  system for convolutional neural networks. In: Proceedings SysML (2019)

\end{thebibliography}

\end{document}